\documentclass[sigconf]{acmart}

\usepackage{algorithm}
\usepackage{algorithmic}
\usepackage{amsmath}
\usepackage{multirow}
\usepackage{bbding}
\usepackage{colortbl}
\definecolor{c1}{HTML}{FF8C00}

\AtBeginDocument{%
  }

\setcopyright{acmlicensed}
\copyrightyear{2026}
\acmYear{2026}
\setcopyright{cc}
\setcctype{by}
\acmConference[KDD '26]{Proceedings of the 32nd ACM SIGKDD Conference on Knowledge Discovery and Data Mining V.2}{August 09--13, 2026}{Jeju Island, Republic of Korea}
\acmBooktitle{Proceedings of the 32nd ACM SIGKDD Conference on Knowledge Discovery and Data Mining V.2 (KDD '26), August 09--13, 2026, Jeju Island, Republic of Korea}
\acmDOI{10.1145/3770855.3817694}
\acmISBN{979-8-4007-2259-2/2026/08}

\begin{document}

\title{Cross-Domain Molecular Relational Learning: Leveraging Chemical Structure-Activity Analysi}

\author{Peiliang Zhang}
\email{cheungbl@ieee.org}
\affiliation{%
  \institution{Wuhan University of Technology}
  \city{Wuhan}
  \state{Hubei}
  \country{China}
}
\affiliation{%
  \institution{Yonsei University}
  \city{Seoul}
  \country{Republic of Korea}
}

\author{Jingling Yuan}
\authornote{Corresponding authors.}
\email{yjl@whut.edu.cn}
\affiliation{%
  \institution{Hubei Key Laboratory of Transportation Internet of Things}
  \institution{State Key Laboratory of Silicate Materials for Architectures}
  \institution{Wuhan University of Technology}
  \city{Wuhan}
  \state{Hubei}
  \country{China}
}

\author{Shiqing Wu}
\email{sqwu@cityu.edu.mo}
\affiliation{%
  \institution{City University of Macau}
  \city{Macau SAR}
  \country{China}
}

\author{Mengqing Hu}
\email{humengqing3@gmail.com}
\affiliation{%
  \institution{Kyung Hee University}
  \city{Seoul}
  \country{Republic of Korea}
}

\author{Chao Che}
\email{chechao@gmail.com}
\authornotemark[1]
\affiliation{%
  \institution{Dalian University}
  \city{Dalian}
  \state{Liaoning}
  \country{China}
}

\author{Yongjun Zhu}
\email{zhu@yonsei.ac.kr}
\affiliation{%
  \institution{Yonsei University}
  \city{Seoul}
  \country{Republic of Korea}
}

\author{Lin Li}
\email{cathylilin@whut.edu.cn}
\affiliation{%
  \institution{Wuhan University of Technology}
  \city{Wuhan}
  \state{Hubei}
  \country{China}
}

\renewcommand{\shortauthors}{Zhang et al.}

\begin{abstract}
Recent advances in molecular representation integrates molecular topological and visual modalities, opening new avenues for precise Molecular Relational Learning (MRL). Existing MRL methods focus on intra-domain modeling, and their inherent domain-closed effect limits applicability to molecular science, particularly in elucidating cross-domain interaction mechanisms. Consequently, the imperative for Cross-Domain Molecular Relational Learning has become increasingly pressing.
Benefiting from structure-activity analysis, we propose the Domain Adversarial Training Network with Structural-Semantic \textbf{Trans}fer \textbf{Dis}crepancy (DisTrans) to optimize cross-domain adaptive representation for molecular structures and visual images. 1) We employ the gradient reversal strategy based on substructure topological discrepancies between domains to learn the domain dependence of molecular structures. This strategy guides the model to adapt to the structural adjacency patterns in the target domain, generating domain-separable structural representations. 2) We apply the cross-domain representation guidance mechanism to align the functional-group semantic information between the source and target domains, learning cross-domain consistency information. 
The experimental results in two typical cross-domain strategies demonstrate that DisTrans outperforms 16 baseline methods, maintaining satisfactory performance even under pronounced inter-domain discrepancy.
\end{abstract}





\received{20 February 2007}
\received[revised]{12 March 2009}
\received[accepted]{5 June 2009}

\maketitle

\section{Introduction}
\label{Introduction}
With the advancement of AI for Science, molecular representation is evolving from traditional structural analysis toward multi-modal semantic understanding~\cite{lee2023cgibICML,zhang2026kdd,xiang2024molecular}. Cross-modal representation methods that integrate molecular structures and images enable models to simultaneously capture molecular topological information and visual semantic priors, achieving a critical leap from structural analysis to semantic comprehension \cite{xu2025deep,chi,he2025imageddi}. This paradigm provides a new research avenue for Molecular Relational Learning (MRL) and establishes a more generalizable representational foundation for downstream applications in molecular science \cite{zhang2025motif,zhang2026prototype}.

The core of MRL lies in molecular representation, which is primarily achieved through molecular structures or molecular images \cite{zhang2025IJCAI}. Researchers typically employ Graph Neural Networks to learn the topological information of molecular structures \cite{lee2023shiftKDD,zhang2025iterative}. Meanwhile, semantic prior information in molecular images (such as functional group types) is extracted via visual models to complement topological representations \cite{zengijcai,ganguly2025merge}.
Overall, current MRL research primarily focuses on intra-domain molecular representations \cite{zhang2025IJCAI,du2024mmgnnIJCAI}. 
However, structural discrepancies across molecular domains render single-domain closed-set molecular representation paradigms ineffective for representing cross-domain molecules. Therefore, advancing Cross-Domain Molecular Relational Learning (CDMRL) is a critical driving force for enabling in-depth analysis of multi-source molecular mechanisms.

\begin{figure}[htpb]
\centering
\includegraphics[width=0.9\columnwidth]{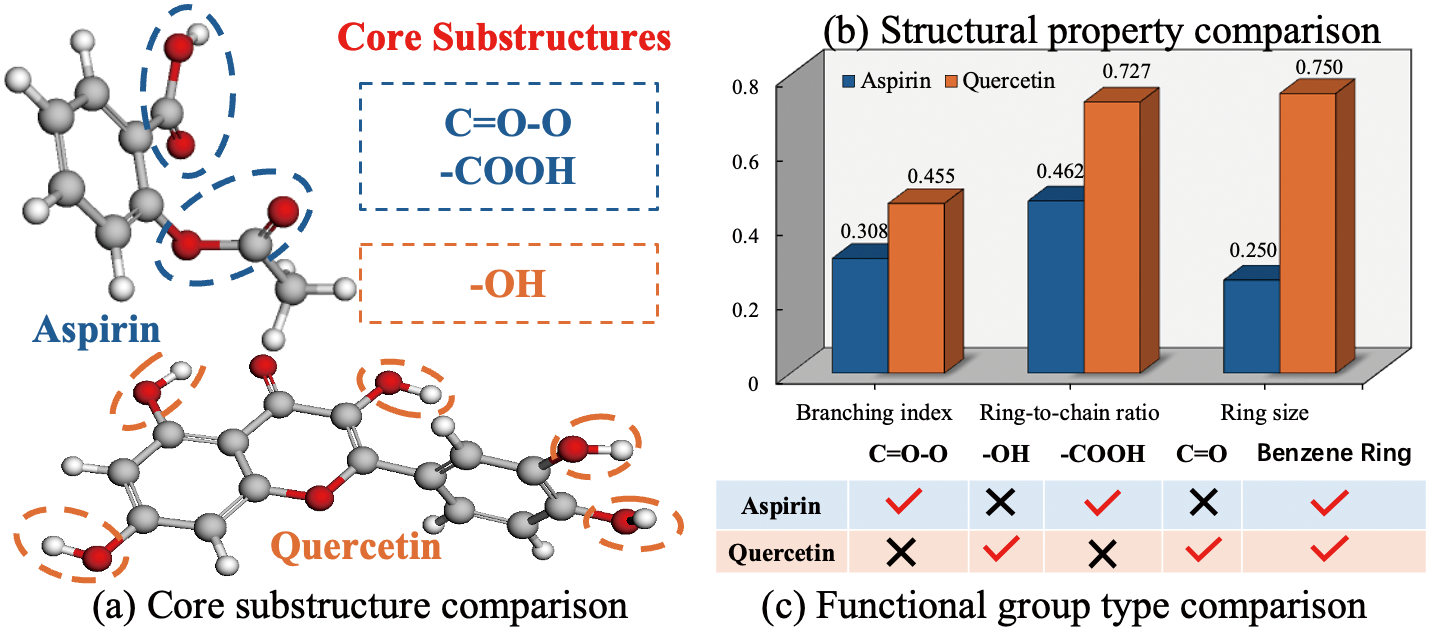}
\caption{The comparison of structural differences between Aspirin (Drug) and Quercetin (Herb).}
\label{motivation}
\end{figure}
Cross-domain learning offers a viable solution to CDMRL, yielding significant results, including domain adversarial learning, domain generalization, and domain transfer \cite{gao2025disrt,sepahvand2025selective}.
Molecular image semantic information (such as functional group and chemical bond types) typically derives from stable chemical prior knowledge, thereby exhibiting strong consistency across domains \cite{zengijcai,he2025imageddi}. Similarity-based cross-domain learning mitigates prior knowledge representation biases by guiding image semantic alignment.
However, the domain dependence of molecular structures poses new challenges for similarity-based transfer paradigms. Molecular structures vary significantly across domains due to inherent domain characteristics \cite{li2025dual,zhang2024key}. As shown in Fig.~\ref{motivation}, drugs are predominantly small and compact, while herbs typically exhibit complex natural scaffolds and cyclic structures. The difference in the number of hydroxyl groups (–OH) between Aspirin and Quercetin directly accounts for their striking disparity in solubility and active site binding modes. These observations indicate that molecular topologies demonstrate significant domain specificity, meaning that molecular topological representations are highly domain-dependent \cite{jiang2025bi,10436119}. The similarity-based cross-domain paradigms primarily focus on feature consistency between domains while neglecting the domain-specific characteristics of molecular structures \cite{xu2025unraveling}. This limitation results in inadequate capacity for capturing domain-specific topological representations in CDMRL.
Therefore, the core challenge in CDMRL lies in synergistically transferring domain-specific structural representations and domain-invariant semantic information.

To address the aforementioned challenges, we first theoretically demonstrate that structure-activity shifts between domains positively correlate with performance degradation in CDMRL, with structural shifts being the dominant factor. Benefiting from chemical structure-activity analysis, we propose the Domain Adversarial Training Network with Structural-Semantic \textbf{Trans}fer \textbf{Dis}crepancy (DisTrans) to enable accurate CDMRL. \textit{DisTrans differentially transfers molecular structure and image representations by maximizing topological representation domain differences while enhancing cross-domain semantic similarity.} First, we design a cross-modal molecular representation as the feature extractor for DisTrans to capture image semantics and structural representations. (1) Within the domain adversarial training, we employ the gradient reversal strategy based on substructure topological discrepancies between domains to drive the adaptation of the model to target-domain structural adjacency patterns, generating domain-dependent structural representations. (2) We apply the cross-domain representation guidance mechanism to align the functional-group semantic information between source and target domains, learning cross-domain consistency information. Finally, we design the mutual information shift-aware fusion mechanism to integrate domain-dependent structural representations with functional-group semantic information dynamically. Experimental results demonstrate that DisTrans outperforms baselines.
\begin{itemize}
    \item We propose a novel DisTrans transfer learning framework that breaks from the conventional similarity-based transfer paradigm to capture domain-specific structural representations. To the best of our knowledge, this is the first work to systematically explore CDMRL.
    \item With theoretical justification, we provide formal certification for DisTrans optimization to enable differentiated transfer of topological and semantic representations through domain-specific enhancement and semantic alignment.
    \item DisTrans achieves performance improvements ranging from 2.29\% to 12.93\% across two cross-domain strategies for three molecular types, maintaining satisfactory performance even under pronounced inter-domain discrepancies.
\end{itemize}

\section{Related Work}
\label{Related Work}

\textbf{Molecular Relational Learning (MRL).} MRL aims to uncover functional or structural interaction between molecules, providing intelligent support for molecular science research \cite{zhang2025iterative,lee2023cgibICML}. Current research in MRL predominantly focuses on learning molecular structural features, primarily through graph-based and image-based methods \cite{zhang2025motif}. The former typically represents molecules as undirected graphs and utilizes Graph Neural Networks to capture structural representation \cite{ren2025predicting}. The latter focuses on extracting chemically relevant prior knowledge from molecular images to complement topological structures for comprehensive representation learning \cite{zengijcai,wang2025image}. Cross-modal molecular representation learning has become the mainstream MRL approach.

\textbf{Cross-Domain Learning.} 
Cross-domain learning aims to address knowledge distribution shifts between source and target domains \cite{xu2025deep}, with tasks primarily including domain adaptation, multi-source domain alignment, and domain generalization \cite{peng2024dual}. Researchers have proposed various strategies, such as domain-invariant representation learning, domain adversarial training, domain reconstruction, and domain label-guided optimization, to mitigate domain distribution discrepancies in Computer Vision and Natural Language Processing \cite{xu2025deep}. Notably, similarity between domains has been widely recognized as a key enabler for successful knowledge transfer \cite{xu2025unraveling}.

Although similarity-based cross-domain paradigms have achieved significant results, they ignore the domain dependency of molecular structures. Therefore, we incorporate chemical structure-activity analysis into DisTrans to achieve differentiated transfer of structural and image representations.

\section{CDMRL}
\label{Cross-Domain Molecular Relational Learning}
In this section, we formally define CDMRL and subsequently provide the constraint objective for CDMRL.

\textbf{CDMRL.}
The core objective of MRL is to learn a mapping function $f_\theta$ between interaction relationships $Y_{ij}$ and molecular pairs $(M_i, M_j)$. Drawing inspiration from cross-domain learning in Computer Vision, we extend it to define CDMRL.
CDMRL aims to learn an MRL model $f_\theta^{s}$ in the source domain $\mathcal{D}_s$, and generalize it to the target domain $\mathcal{D}_t$, while maintaining the model's predictive performance. Formally, $f_\theta^{s}(M_i^t,M_j^t) \!\! \rightarrow \!\! Y_{ij}^t$.
We adopt both molecular images $\mathcal{F}$ and molecular graphs $\mathcal{G}$ as the original representations of molecules. The molecular images $\mathcal{F}_M \!\! \in \!\! \mathbb{R}^{H \times W \times C}\!\!\!=\!\!\!\{\mathcal{P}_1,\mathcal{P}_2,\cdots ,\mathcal{P}_L\}$ are composed of $L$ fixed-size patches $\mathcal{P}$, where $H$, $W$, and $C$ denote the height, width, and channels of the image, respectively. The molecular structures $\mathcal{G}_M\!\!=\!\!\{\mathcal{V},\!\mathcal{E}\}$ are represented as an undirected graph. $\mathcal{V}\!\!\!=\!\!\!\{\mathcal{V}_1,\!\mathcal{V}_2,\!\cdots, \!\mathcal{V}_N\!\}$ denotes the set of $N$ atoms, and $\mathcal{E}$ is the adjacency matrix representing chemical bonds.

\begin{figure*}[t]
\centering
\includegraphics[width=\textwidth]{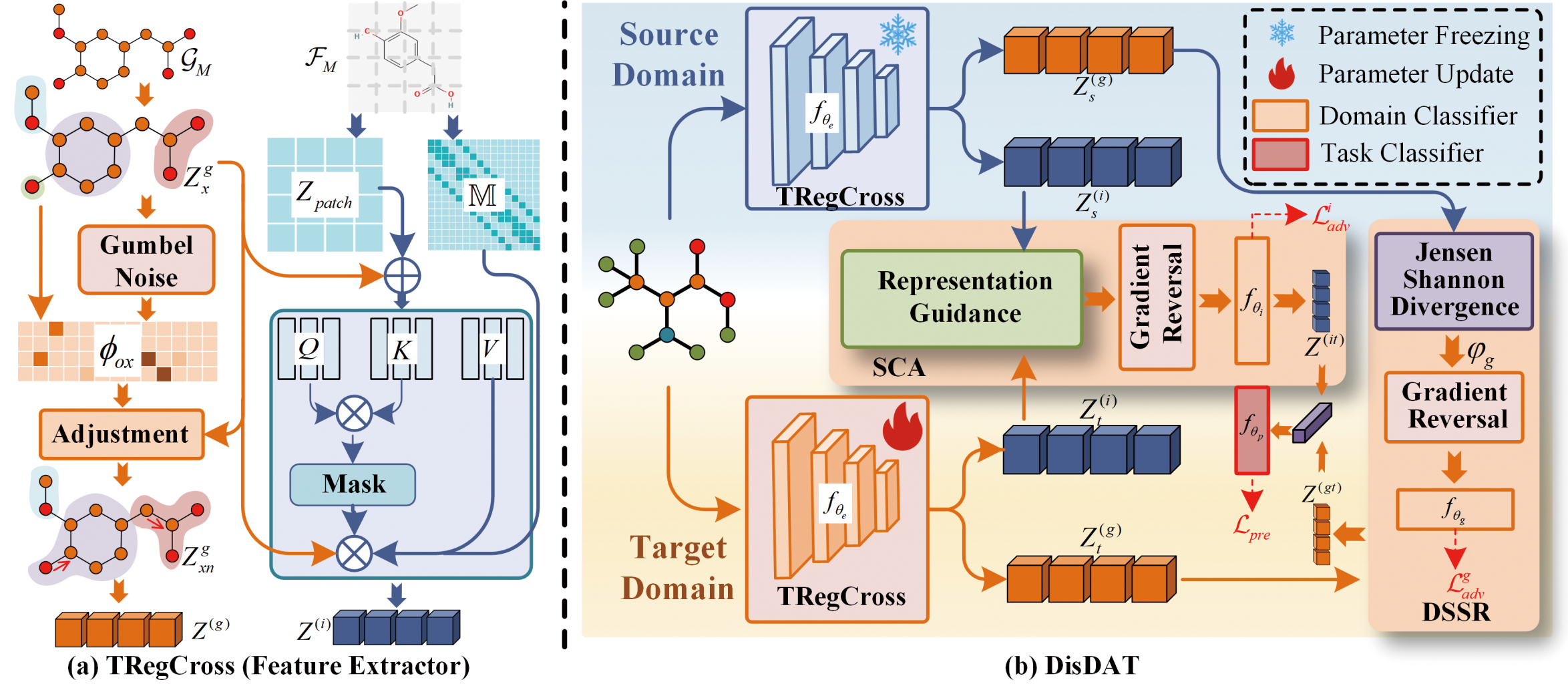} 
\caption{The framework of DisTrans. (a) TRegCross serves as the feature extractor to capture molecular representation. (b) DisDAT updates gradient representations by JSD-based substructure topological discrepancies to capture domain-specific topological features, while combining cross-domain representation guidance to align semantic information between domains.}
\label{MODEL}
\end{figure*}

\textbf{CDMRL Constraint.}
We explore the factors affecting performance degradation in CDMRL by domain adaptation theory \cite{xu2025unraveling}.
With cross-domain structure-activity shift represented by mutual information, we establish the generalization error theory for CDMRL.

\textit{\textbf{Theorem 1 (Cross-Domain Generalization Error).}
Let $Z^{(g)}\!\!=\!\!f_\theta(\mathcal{G})$ and $Z^{(i)}\!\!=\!\!f_\theta(\mathcal{F})$ denote the topological and image representations of molecule learned by the model, $Z\!\!=\!\!(Z^{(g)}, Z^{(i)})$. The representation shift $\Delta\mathcal{I}$ between source domain $\mathcal{D}_s$ and target domain $\mathcal{D}_t$ is $\Delta\mathcal{I}\!\!=\!\mathcal{I}\!(Z_s;Y_s) \!-\! \mathcal{I}\!(Z_t;Y_t)\!=\!\Delta\mathcal{I}^{(g)}\!\!+\!\Delta\mathcal{I}^{(i|g)}$. Then, there exists a real number $\xi\!>\!0$ such that the prediction error $\Delta\varepsilon$, the structure shift $\Delta\mathcal{I}^{(g)}$, and the image shift $\Delta\mathcal{I}^{(i|g)}$ between domains satisfy:}
\begin{equation}
    \Delta\varepsilon \geq \xi (\Delta\mathcal{I}^{(g)}+\Delta\mathcal{I}^{(i|g)}),\quad \Delta\mathcal{I}^{(g)} \gg \Delta\mathcal{I}^{(i|g)}
    \label{the1}
\end{equation}

Theorem 1 indicates that the predictive performance of CDMRL is positively correlated with the structure-activity shift between molecular representations across domains, with structural shift being the dominant factor. The proof of Theorem 1 is provided in Appendix \ref{The Proof of Theorem 1}.
With the above conclusions and chemical structure-activity analysis, we formalize the core optimization objective of CDMRL as task discriminability of molecular representation $Z$, domain discriminability of topological representation $Z^{(g)}$, and domain indistinguishability of semantic information $Z^{(i|g)}$.

\textit{\textbf{Definition 1 (CDMRL Constraint).}
Let $\mathcal{I}\!(Z;Y)$, $\mathcal{I}\!(Z^{(g)};A)$, and $\mathcal{I}\!(Z^{(i|g)};A)$ are the mutual information between $Z$ and task labels $Y$, $Z^{(g)}$ and domain labels $A$, and $Z^{(i|g)}$ and domain labels $A$, respectively. Then, the core parameter $\theta$ in CDMRL satisfies the constraints:}
\begin{equation}
\min_\theta -\mathcal{I}(Z;Y)-\lambda_g \mathcal{I}(Z^{(g)};A)+\lambda_i \mathcal{I}(Z^{(i|g)};A)
    \label{aim}
\end{equation}
\textit{$\lambda_g$ and $\lambda_i$ represent the adaptive weighting coefficients for target domain's graph and image modality, respectively.}

Evidently, the upper bound of $\mathcal{I}(Z;Y)$ can be optimized by minimizing task loss. $\mathcal{I}(Z^{(g)};A)$ emphasizes domain specificity of topological representations, $\mathcal{I}(Z^{(i|g)};A)$ aims to generate consistent semantic information.

\section{DisTrans Methodology}
\label{Methodology}
Motivated by Theorem 1 and chemical structure–activity analysis, we propose DisTrans, which achieves CDMRL by perceiving modality transfer discrepancies, as shown in Fig.~\ref{MODEL}. With the Topology-Regulated Cross-Modal Molecular Representation (TRegCross) in Fig.~\ref{MODEL}(a), we design the Domain Adversarial Training with Structural–Semantic Discrepancy (DisDAT) to jointly learn domain-specific topological representations and domain-consistent semantic information, depicted in Fig.~\ref{MODEL}(b).

\subsection{TRegCross}
Considering the deviation between semantic and structural representation caused by conformational variations (changes in functional group orientation), we propose TRegCross for consistent representations. 
We guide image semantics via a chemical structure mask attention to learn functional group prior knowledge.

\textbf{Structural Topological Representation.} 
To address the challenge of dynamically optimizing substructure representations based on rule-based partitioning, we introduce adjacent low-noise atom injection to adaptively optimize substructure topology while preserving chemical information.
 
Following \cite{wang2024advancing}, we divide the molecular graph into substructures $\{\mathcal{S}_1,\mathcal{S}_2,\cdots,\mathcal{S}_X\}$ based on the bond-breaking rules in RDKit \cite{brown2015silico}, where $X$ depends on the bond rules and molecular sequence. For each substructure $\mathcal{S}_x$, we utilize GINE-based \cite{Hu2020Strategies} adjacency-weighted aggregation to capture the original embedding $Z^{g}_x$.
\begin{equation}
    Z^{g}_x=\gamma \sum \nolimits_{\mathcal{V}_x \in \mathcal{S}_x}\!\!Z^{\mathcal{V}}(\mathcal{V}_x)+\frac{1-\gamma}{|\mathcal{S}_x|} \sum \nolimits_{\mathcal{V}_{o} \neq \mathcal{V}_{x}}\!\! Z^{\mathcal{V}}(\mathcal{V}_{o})
    \label{subembedding}
\end{equation}
Where $\gamma$ is the residual factor, and $Z^{\mathcal{V}}(\mathcal{V}_x)$ represents the atom $\mathcal{V}_x$ embedding learned by  GINE encoder.

Since $Z^{g}_x$ primarily emphasizes the topological representation within substructure, neglecting the influence of adjacent atoms, we propose a Gumbel-Softmax-based adjacent information injection mechanism to optimize substructure representations. For the substructure $\mathcal{S}_x$, we compute the affiliation probability $\eta = MLP(Z^{g}_x) \cdot \sigma(Z^{\mathcal{V}}(\mathcal{V}_{adj})^\top Z^{g}_x)$ based on the embedding $Z^{\mathcal{V}}(\mathcal{V}_{adj})$ of adjacent atoms and $Z^{g}_x$. The noise distribution $\phi_{ox}$ is calculated with $\eta$.
\begin{equation}
    \phi_{ox}=\frac{\exp ((\log \eta+Gum_{x})/\tau)}{\sum \nolimits^{X}_{o=1}\exp ((\log \eta+Gum_{o})/\tau)}
    \label{phi}
\end{equation}
Here, $Gum_{x} \sim \mathrm{Gumbel}(0,1)$ is noise sampled from the Gumbel distribution. Unlike continuous perturbation noise, the differentiable categorical sampling of Gumbel noise facilitates the selection of salient atoms. $\tau$ is the temperature coefficient. Finally, we select the embeddings of $Q$ low-noise atoms based on $\phi_{ox}$ and fuse them with $Z^{g}_x$ to generate the embedding representation $Z^{g}_{xn}$ of $\mathcal{S}_x$.
\begin{equation}
    Z^{g}_{xn}=\sum \nolimits_{q=1}^{Q} \phi_{qx} \cdot Z^{(\mathcal{V})}(\mathcal{V}_{q}) || Z^{g}_x
    \label{zsnew}
\end{equation}

\textbf{Image Semantic Information Learning.}
To achieve consistent representations between structure and images, we propose a topology-guided chemical structure masked attention mechanism to mitigate semantic bias caused by image conformation sensitivity.

Specifically, we utilize RDKit \cite{brown2015silico} to divide molecular images into fixed-size patches and construct the chemical structure mask matrix $\mathbb{M}$ based on substructure integrity and patch adjacency. Finally, we construct a local semantic-enhanced attention mechanism $ATT_{se}$ with $\mathbb{M}$ in the ViT framework \cite{dosovitskiy2020image}.
\begin{equation}
    \mathbb{M}(\mathcal{P}_l,\!\mathcal{P}_k)=\begin{cases}0 & \text{if} \!\!\quad \mathcal{P}_l \!\!\quad\text{adjacent}\quad \!\!\mathcal{P}_k \lor\{\!\mathcal{P}_l,\!\mathcal{P}_k\!\} \!\!\in \!\!\mathcal{S}_x\\ -\infty & \text{otherwise}\end{cases}
    \label{M}
\end{equation}
\begin{equation}
    ATT_{se}\!=\!\sigma \!\left(\frac{QK^{\mathsf{T}}}{\sqrt{d}}\!+\!\mathbb{M}(\mathcal{P}_l,\mathcal{P}_k)\cdot \!Z^{g}_x \right)V, \!\!\quad \text{s.t.} \!\!\quad l,k\le \! L
    \label{ATTV}
\end{equation}
The adjacency information of $\mathcal{P}_l$ is computed based on the coordinate mapping generated by RDKit \cite{brown2015silico}. $Q$, $K$, $V$ and $d$ are the same as in ViT. 
The input of ViT consists of patch embeddings ${Z}_{patch}$ represented by pretrained ImageMol \cite{zeng2022accurate} and $Z^{g}_x$. 
After the ViT framework, we obtain the image semantic embedding $Z^{(i)}(M)$ of molecule $M$.

\subsection{DisDAT}
Benefiting from chemical structure-activity analysis, we propose DisDAT consisting of feature extractor $f_{\theta_{e}}(\cdot)$, structure domain classifier $f_{\theta_{g}}(\cdot)$, image domain classifier $f_{\theta_{i}}(\cdot)$, and task domain classifier $f_{\theta_{p}}(\cdot)$. 
DisDAT differentially transfers molecular structure and image representations by maximizing topological representation domain differences and cross-domain semantic similarity.

\textbf{Domain-Specific Structural Representation (DSSR).} 
The primary obstacle to molecular topological representation transfer is the domain dependence of molecular structures. Therefore, structural representation transfer aims to generate domain-specific structural representation by capturing molecular structure differences between domains.

To achieve this objective, we introduce the substructure representation discrepancies-based gradient reversal strategy to guide the model in adapting to the target domain’s structural adjacency patterns and learning domain-specific structural representations. Specifically, we quantify the substructure representation topological discrepancy $\varphi_{g}$ between domains using the Jensen-Shannon Divergence (JSD) \cite{jsd-tkde} and employ it as the gradient scaling factor to emphasize domain-specific structural learning. During backpropagation, the gradient reversal mechanism captures the gradient $g_{n+1}$ from the subsequent layer and passes it to the previous layer $g_{n}$. Formally, $g_{n} = -\varphi_{g} \cdot g_{n+1}$.

\begin{equation}
    \varphi_{g}=\!\mathbb{E}_{Z \sim \mathbb{P}_s}\!\left[\log \!
    \frac{2 \mathbb{P}_s(Z)}{\mathbb{P}_s(\!Z)\!+\!\mathbb{P}_t(\!Z)}\right]\!+\!\mathbb{E}_{Z \sim \mathbb{P}_t}\!\left[\log \!
    \frac{2 \mathbb{P}_t(Z)}{\mathbb{P}_s(Z)\!+\!\mathbb{P}_t(Z)}\right]
    \label{varphig}
\end{equation}
Here, $\mathbb{P}(Z)\!\!=\!\!\mathbb{E}_{Z^g_{xn} \sim \{Z^g_{1n},\cdots,Z^g_{Xn}\}}\![\delta(\!(\sum \nolimits_{x=1}^X \!Z^g_{xn})\!\!-\!\!Z^g_{xn})]$ denotes the structural representation distribution calculated by the Dirac function $\delta(\cdot)$.
In the structural domain classifier, the structural domain classification loss function $\mathcal{L}^{g}_{adv}$ is defined as the joint representation of the structural representations $Z^{(g)}$ generated by  $f_{\theta_{e}}(\cdot)$ and $f_{\theta_{g}}(\cdot)$.
\begin{equation}
    \mathcal{L}_{adv}^{g}\!=\!-\mathbb{E}_{Z^{(g)} \sim \mathbb{P}_s}\!\left[ \log(1\!-\!f_{\theta_{g}}(Z^{(g)})) \right]\!-\!\mathbb{E}_{Z^{(g)} \sim \mathbb{P}_t}\!\left[ \log f_{\theta_{g}}(Z^{(g)}) \right]
    \label{ladvg}
\end{equation}
Where $f_{\theta_{g}}(Z^{(g)})$ indicates the classifier's predicted probability of the sample originating from the source domain. By minimizing $\mathcal{L}^{g}_{adv}$, \textit{i.e.}, maximizing the structural representation distribution between the source and target domains, DisTrans is encouraged to learn domain-specific topological representations $Z^{(gt)}$.

\textbf{Semantic Consistency Alignment (SCA).}
The semantic information in molecular images serves as stable chemical prior knowledge and exhibits strong consistency across domains. Consequently, semantic information transfer aims to generate consistent information by maximizing cross-domain representational similarity.

Specifically, we incorporate semantic similarity between domains into the image domain classifier via a similarity-guided representation transfer strategy. During backpropagation, negative gradients are applied to update the representation gradients. The image domain classifier optimizes $f_{\theta_{i}}(\cdot)$ to align inter-domain representation gradients, thereby generating domain-independent prior semantic information.
Based on the semantic representations $Z^{(i)}$ and the source and target feature distributions $\mathbb{P}_s$ and $\mathbb{P}_t$, the image domain classification loss function $\mathcal{L}_{adv}^{i}$ is defined as:
\begin{equation}
\mathcal{L}_{adv}^{i}=-\mathbb{E}_{Z^{(i)} \sim \mathbb{P}_s \cup \mathbb{P}_t}[\log f_{\theta_{i}}(Z^{(i)})]
    \label{ladvi}
\end{equation}

By minimizing $\mathcal{L}_{adv}^{i}$, \textit{i.e.}, maximizing the semantic similarity between the source and target domains, DisTrans achieves consistent cross-domain representation $Z^{(it)}$ of image semantics.

\textbf{Mutual Information Shift-Aware Representation Fusion.}
Theorem 1 indicates that structure shift is the dominant factor affecting the prediction performance of CDMRL. Therefore, we adopt the contribution of modality-specific mutual information shifts as the fusion measure to generate the cross-domain molecular representation $Z_M$ by integrating $Z^{(gt)}$ and $Z^{(it)}$.
\begin{equation}
    Z_M= \frac{\Delta \mathcal{I}^{(g)}}{\Delta \mathcal{I}}\cdot Z^{(gt)}+\frac{\Delta \mathcal{I}^{(i|g)}}{\Delta \mathcal{I}}\cdot Z^{(it)}
    \label{fusion}
\end{equation}
$\Delta \mathcal{I}$ denotes the cross-domain mutual information shift estimated by Mutual Information Neural Estimation \cite{belghazi2018mutual}. Finally, the task prediction loss $\mathcal{L}_{pre}$ based on $f_{\theta_{p}}(\cdot)$ and $(M_i,M_j,Y_{ij})$ is defined as:
\begin{equation}
    \mathcal{L}_{pre}=-\mathbb{E}_{(Z_{M_i},Z_{M_j},Y_{ij})\sim D_t}[\log f_{\theta_{p}}(Z_{M_i},Z_{M_j})|Y_{ij}]
    \label{pred}
\end{equation}
We set $\mathcal{L}_{pre}$ as Mean Absolute Error Loss (regression task) or Cross-Entropy Loss (classification task) with task type.

\subsection{Model Optimization and Analysis}
\textbf{Model Optimization.}
Aligning with the CDMRL constraint in Eq (\ref{aim}), we design the loss function $\mathcal{L}oss$ with task prediction loss $\mathcal{L}_{pre}$ and domain label prediction loss $\mathcal{L}_{adv}$.
\begin{equation}
\mathcal{L}oss=\mathcal{L}_{pre}+\lambda_g\mathcal{L}_{adv}^{g}+\lambda_i\mathcal{L}_{adv}^{i}
    \label{losssum}
\end{equation}
Where $\lambda_g$ and $\lambda_i$ are set consistently with Eq (\ref{aim}).

\textbf{Model Analysis.}
DisTrans is trained by optimizing TRegCross within DisDAT. The pseudocode is in Algorithm \ref{alg:DisTrans}. The computational complexity of TRegCross primarily arises from the iterative optimization of GINE and ViT, with complexities of $\mathcal{O}(Lg \!\cdot\! N \!\cdot\! Fg^2)$ and $\mathcal{O}(Lv \!\cdot\! L^2 \!\cdot\! Fp)$, respectively, where $Lg$ and $Lv$ denote the number of GINE and ViT layers, and $Fg$ and $Fp$ are the feature dimensions of atoms and patches. The cross-domain optimization's computational complexity $\mathcal{O}(B \cdot Fc)$ depends on the batch size $B$ and feature dimension $Fc$. Since most cross-domain operations are linear, TRegCross primarily determines the overall computational complexity of DisTrans. Therefore, DisTrans does not significantly increase the computational complexity compared to non-cross-domain methods.

\begin{algorithm}[t]
    \caption{DisTrans}
    \label{alg:DisTrans}
    \textbf{Input}: The sampled source domain data $(M_i^s,M_j^s,Y_{ij})$, target domain data $(M_i^t,M_j^t,Y_{ij})$, feature extractor $f_{\theta_{e}}(\cdot)$, structural domain label mapping function $f_{\theta_{g}}(\cdot)$, image domain label mapping function $f_{\theta_{i}}(\cdot)$, task label mapping function $f_{\theta_{p}}(\cdot)$, learning rate $\mu$, weighting coefficients $\lambda_{g}$ and $\lambda_{i}$, and the number of training iterations $T$.\\
    \textbf{Parameters}: $\theta_{e}$, $\theta_{g}$, $\theta_{i}$, $\theta_{p}$.\\
    \textbf{Output}: Optimal parameters.
    \begin{algorithmic}[1] 
        \STATE Initialized parameters $\theta_{e}$, $\theta_{g}$, $\theta_{i}$, $\theta_{p}$.
        \FOR{$t_1=1$ to $T$}{
        \STATE \textit{\textbf{/*** Forward Pass ***/}}
        \STATE Generate source domain features $Z^s=(Z_s^{(g)},Z_s^{(i)})$ and target domain features $Z^t=(Z_t^{(g)},Z_t^{(i)})$ based on $Z=f_{\theta_{e}}(M)$;
        \STATE Compute domain labels $\hat{Y}_{g}$ and $\hat{Y}_{i}$ by $\hat{Y}_{g}=f_{\theta_{g}}(Z^{(g)})$ and $\hat{Y}_{i}=f_{\theta_{i}}(Z^{(i)})$;
        \STATE Compute task label $\hat{Y}_{ij}$ by $\hat{Y}_{ij}=f_{\theta_{p}}(Z_{M_i},Z_{M_j})$; 
        \STATE Compute $\mathcal{L}_{adv}^{g}$, $\mathcal{L}_{adv}^{i}$, and $\mathcal{L}_{pre}$ based on Eqs (\ref{ladvg}), (\ref{ladvi}), and (\ref{pred});
        \IF{$(M_i,M_j) \in \mathcal{D}_t$}
        \STATE \textit{\textbf{/*** Gradient Reversal and Parameter Updates ***/}}
        \STATE $\theta_{e} \!\!\gets\! \theta_{e} \!\!- \!\!\mu \!\left(\! \triangledown\!_{\theta_{e}} \!\mathcal{L}_{pre}\!+\! \lambda_g \!\triangledown\!_{\theta_{e}}\! \mathcal{L}_{adv}^g\!+\! \lambda_i \!\triangledown\!_{\theta_{e}} \!\mathcal{L}_{adv}^i\right)$;
        \STATE $\theta_{g} \gets \theta_{g} - \mu  \lambda_g   \varphi_{g} \triangledown_{\theta_{g}} \mathcal{L}_{adv}^g$;
        \STATE $\theta_{i} \gets \theta_{i} - \mu \lambda_i \triangledown_{\theta_{i}} \mathcal{L}_{adv}^i$;
        \STATE $\theta_{p} \gets \theta_{p} - \mu  \triangledown_{\theta_{p}}\mathcal{L}_{pre}$;
        \ELSE
        \STATE $t_1 \gets t_1+1$;
        \ENDIF
        }
        \ENDFOR
    \end{algorithmic}
\end{algorithm}

\section{Experiments}
\label{Experiments}
This section addresses four core research questions by experimental evaluation and analysis:
\textbf{RQ1:} The advantages of DisTrans over non-CDMRL methods in data requirements and predictive performance.
\textbf{RQ2:} The transfer preferences of different types of molecular data and their impact on cross-domain representation learning.
\textbf{RQ3:} The adaptability and generalization of the structure-semantic transfer discrepancy-based cross-domain strategy.
\textbf{RQ4:} Whether the experimental results can be effectively supported and interpreted by chemical domain knowledge.

\begin{table*}[t]
\caption{The dataset details and data settings in CDMRL (Source Domain $\mapsto$ Target Domain).}
\small
\label{dataset}
\centering
\setlength{\tabcolsep}{0.75mm}{
\begin{tabular}{cccc|ccccc|ccccc|ccccc}
\toprule
\multirow{2}{*}{Dataset} & \multirow{2}{*}{\#Pairs} & \multirow{2}{*}{Domain} &  \multirow{2}{*}{Task}& \multicolumn{5}{c|}{IDHT} & \multicolumn{5}{c|}{CDT (Drug $\mapsto$ Herb)} & \multicolumn{5}{c}{CDT (Herb $\mapsto$ Drug)}\\ 
 &  &  &  & $\mathcal{D}_s$ & $\mathcal{D}_t$ & \#$\mathcal{D}_{Tra}^{30\%}$ & \#$\mathcal{D}_{Tra}^{50\%}$ & \#$\mathcal{D}_{Tra}^{70\%}$ & $\mathcal{D}_s$ & $\mathcal{D}_t$ & \#$\mathcal{D}_{Tra}^{30\%}$ & \#$\mathcal{D}_{Tra}^{50\%}$ & \#$\mathcal{D}_{Tra}^{70\%}$ &  $\mathcal{D}_s$ & $\mathcal{D}_t$ & \#$\mathcal{D}_{Tra}^{30\%}$ & \#$\mathcal{D}_{Tra}^{50\%}$ & \#$\mathcal{D}_{Tra}^{70\%}$ \\
 \midrule
 Chromophore \cite{chromophore} & 18141 & Chemical & MI &  & \Checkmark & 9500 & 13129 & 16757 &  &  &  &  &  &  &  &  &  &    \\
 MNSol \cite{mnsol} & 2275 & Chemical & MI &  & \Checkmark & 4741 & 5196 & 5651 &  &  &  &  &  &  &  &  &  &  \\
 CombiSolv \cite{combisolv} & 10145 & Chemical & MI & \Checkmark &  &  &  &  &  &  &  &  &  &  & &  &  &  \\
 \midrule
 DDInter2.0 \cite{DDInter} & 33500 & Drug & DDI &  & \Checkmark & 26152 & 32852 & 39552 &  &  & &  &  &  & \Checkmark  & 31717 & 38417 & 45117\\
 DrugMap2024 \cite{DrugMAP} & 151950 & Drug & DDI &  & \Checkmark & 61687 & 92077 & 122467 &    &  &   &  &  &  & \Checkmark  & 69412 & 101242 & 133072\\
 ZhangDDI \cite{zhangddi} & 40255 & Drug & DDI & \Checkmark &  &  &  &  &  \Checkmark & &  &  &  &  & \Checkmark  & 33744 & 41794 & 49846 \\
 \midrule
 TCMM \cite{tcmm} & 49506 & Herb & HHI & & \Checkmark & 36519 & 46420 & 56321 & &  \Checkmark & 30954 & 40855 & 50756 &  & &  &  &   \\
 ITCM \cite{itcm} & 54167 & Herb & HHI &  \Checkmark &   & &  &  & &  \Checkmark & 32352 & 43186 & 54019 & \Checkmark &  &  &  &  \\
\bottomrule 
\end{tabular}
}
\end{table*}

\subsection{Experimental Setup and Datasets}
\label{Experimental Setup}
\subsubsection{Dataset}
\label{Dataset}
Following related MRL studies \cite{zhang2025iterative,lee2023cgibICML,lee2023shiftKDD}, we conducted extensive CDMRL experiments on eight datasets involving three molecular types, as shown in Table~\ref{dataset}. We performed solute-solvent free energy prediction for chemical molecules as a proxy for Molecular Mnteraction (MI) prediction using the Chromophore \cite{chromophore}, MNSol \cite{mnsol}, and CombiSolv \cite{combisolv} datasets. For drug molecules, we conducted Drug-Drug Interaction (DDI) prediction on the DDInter2.0 \cite{DDInter}, DrugMap2024 \cite{DrugMAP}, and ZhangDDI \cite{zhangddi} datasets. For herb molecules, Herb-Herb Interaction (HHI) prediction was performed using the TCMM \cite{tcmm} and ITCM \cite{itcm} datasets.
To better simulate cross-domain application scenarios in molecular science, we designed two molecular cross-domain strategies: Intra-Domain Heterogeneous Transfer (IDHT) and Cross-Domain Transfer (CDT). 
\begin{itemize}
    \item IDHT refers to the CDMRL conducted between different datasets within the same molecular domain. Taking the example of chemical molecules, our model optimize the feature extractor on the CombiSolv dataset and then performe cross-domain adaptation on the Chromophore and MNSol datasets. In contrast, the baseline methods are trained and evaluated directly on the Chromophore and MNSol datasets, i.e., their predictive performance is non-cross-domain performance.
    \item CDT refers to CDMRL performed across datasets from different molecular domains. Specifically, our method and the baseline methods are trained on drug molecule domain datasets and tested on datasets from the herb molecule domain.
\end{itemize} 

During TRegCross training, we utilize 60\% of the source-domain data. For cross-domain transfer learning, we employ 40\% of the source domain data alongside fixed proportions (30\%, 50\%, 70\%) of the target domain data for training, reserving the remaining target domain data for testing.
Baseline methods followed the same data configuration for model training and cross-domain evaluation.
Table~\ref{dataset} presents the source domain data selection and the number of positive samples in the target domain of different cross-domain strategies. Negative samples are generated based on molecular scaffold similarity \cite{zhang2025IJCAI}.

\subsubsection{Evaluation Metrics}
\label{Evaluation Metrics}
According to the nature of each task, MI prediction is treated as a regression task, while DDI and HHI predictions are treated as classification tasks. Given the high positive correlation between Root Mean Square Error (RMSE) and Mean Absolute Error (MAE), we adopt RMSE as the evaluation metric for MI prediction. For DDI and HHI prediction tasks, classification performance is evaluated by the Area Under the Receiver Operating Characteristic Curve (AUROC), Accuracy (ACC), F1-score (F1), Precision (Pre), and the Area Under the Precision-Recall Curve (AUPR).

\subsubsection{Training Details}
\label{Training Details}
During domain adversarial training, we froze the feature encoder parameters in the source domain and only optimized the parameter gradients in the target domain. In TRegCross, the GINE and ViT encoders were set to 3 layers. We implemented and trained the DisTrans based on the PyTorch framework on two NVIDIA GeForce RTX 5090 32G GPUs.
The training process is conducted over 200 epochs with a batch size of 64.
The learning rate $\mu$ is selected from $\{0.01,0.005,0.001,0.0001\}$.
The weighting coefficients $\lambda_g$ and $\lambda_i$ are optimized via grid search within $\{0.7, 0.5, 0.3, 0.1, 0.01\}$. The implementation code of DisTrans will be made publicly available upon acceptance of this manuscript.

\subsubsection{Baseline Methods}
\label{Baseline Methods}
To comprehensively evaluate the performance of DisTrans, we selected four graph benchmark models (GCN \cite{gcn}, GAT \cite{gat}, MPNN \cite{mpnn}, and GINE \cite{Hu2020Strategies}), three drug molecular representation models (SSI-DDI \cite{ssiddi}, SADDI \cite{saddi}, and DSN-DDI \cite{dsn}), seven MRL models (CIGIN \cite{cigin}, MIRACLE \cite{miracle}, CMRL \cite{lee2023shiftKDD}, CGIB \cite{lee2023cgibICML}, MMGNN \cite{du2024mmgnnIJCAI}, CausalGIB \cite{zhang2025IJCAI}, and IGIB-ISE \cite{zhang2025iterative}), and tow molecular cross-domain models (Meta-MolNet \cite{10436119}, BIT \cite{zhu2025generalist})  as baseline methods. For MI prediction, we compared the performance of DisTrans with both the graph benchmarks, the MRL models and the molecular cross-domain models. For DDI and HHI prediction tasks, DisTrans was compared with all baseline methods.

\subsection{Overall Performance Analysis (\textbf{RQ1})}
The experimental results are shown in Tables~\ref{IDHT result} and~\ref{CDT result} where the best performance is highlighted in \colorbox{c1!80}{\textbf{bold}}, and the second-best performance is \colorbox{c1!36}{underlined}. $Imp$ is computed according to Eq (\ref{imp}) in Appendix \ref{Overall Performance Analysis}.

\begin{table*}[t]
\caption{The prediction results of DisTrans and baseline methods in Intra-Domain Heterogeneous Transfer (IDHT).}
\label{IDHT result}
\centering
{\fontsize{9pt}{10pt}\selectfont
\setlength{\tabcolsep}{0.8mm}{
\begin{tabular}{l|cc|ccc|ccc|ccc}
\toprule
\multirow{3}{*}{Model} & \multicolumn{2}{c|}{MI ($\mathcal{D}_s$: CombiSolv)} & \multicolumn{6}{c|}{DDI ($\mathcal{D}_s$: ZhangDDI)} & \multicolumn{3}{c}{HHI ($\mathcal{D}_s$: ITCM)}\\
 \cline{2-12}
 & Chromophore & MNSol & \multicolumn{3}{c|}{DDInter2.0} & \multicolumn{3}{c|}{DrugMap2024} & \multicolumn{3}{c}{TCMM}\\
 & RMSE $\downarrow$ & RMSE $\downarrow$ & ACC $\uparrow$ & AUROC $\uparrow$ & AUPR $\uparrow$ & ACC $\uparrow$ & AUROC $\uparrow$ & AUPR $\uparrow$ & ACC $\uparrow$ & AUROC $\uparrow$ & AUPR $\uparrow$\\ 
\midrule
 GCN \cite{gcn} & 31.87$\pm$1.70 & 0.675$\pm$0.021 & 0.6237 & 0.6238 & 0.5727 & 0.8528 & 0.8528 & 0.8883 & 0.6105 & 0.6264 & 0.6356\\
 GAT \cite{gat} & 30.90$\pm$1.01 & 0.731$\pm$0.007 & 0.6246 & 0.6243 & 0.5853 & 0.8571 & 0.8571 & 0.8916 & 0.6352 & 0.6527 & 0.6566\\
 MPNN \cite{mpnn} & 30.17$\pm$0.99 & 0.682$\pm$0.014 & 0.6240 & 0.6240 & 0.5657 & 0.8582 & 0.8582 & 0.8933 & 0.6077 & 0.6324 & 0.6391\\
 GINE \cite{Hu2020Strategies} & 32.31$\pm$0.26 & 0.669$\pm$0.017 & 0.6246 & 0.6244 & 0.5853 & 0.8464 & 0.8464 & 0.8840 & 0.6129 & 0.6346 & 0.6417\\
 SSI-DDI \cite{ssiddi} & - & - & 0.6127 & 0.6627 & 0.6193 & 0.6376 & 0.6934 & 0.6690 & 0.6676 & 0.6918 & 0.7016\\
 SA-DDI \cite{saddi} & - & - & 0.6396 & 0.6729 & 0.6325 & 0.7813 & 0.8545 & 0.8184 & 0.6573 & 0.6845 & 0.6881\\
 DSN-DDI \cite{dsn} & - & - & 0.5948 & 0.5573 & 0.5723 & 0.6819 & 0.6604 & 0.6377 & 0.6245 & 0.6477 & 0.6563\\
 CIGIN \cite{cigin} & 25.09$\pm$0.32 & 0.607$\pm$0.024 & 0.5709 & 0.5709 & 0.5391 & 0.8233 & 0.8233 & 0.7576 & 0.6298 & 0.6455 & 0.6541\\
 MIRACLE \cite{miracle} & 25.75$\pm$0.28 & 0.631$\pm$0.022 & 0.5493 & 0.5570 & 0.5347 & 0.6046 & 0.6480 & 0.6201 & 0.6854 & 0.7129 & 0.7183\\
 CMRL \cite{lee2023shiftKDD} & 24.30$\pm$0.22 & 0.551$\pm$0.017 & 0.6319 & 0.6801 & 0.6400 & \cellcolor{c1!80}\textbf{0.8976} & \cellcolor{c1!36}\underline{0.9548} & \cellcolor{c1!36}\underline{0.9454} &  0.7015 & \cellcolor{c1!36}\underline{0.7183} & 0.7049\\
 CGIB \cite{lee2023cgibICML} & 24.90$\pm$0.35 & \cellcolor{c1!36}\underline{0.538$\pm$0.007} & 0.6304 & 0.6773 & 0.6416 & 0.8843 & 0.9457 & 0.9325 & 0.7085 & 0.7173 & 0.6963\\
 MMGNN \cite{du2024mmgnnIJCAI} & 25.33$\pm$0.43 & 0.546$\pm$0.011 & 0.6015 & 0.6325 & 0.6008 & 0.8522 & 0.8523 & 0.7794 & 0.6821 & 0.6932 & 0.6960\\
 CausalGIB \cite{zhang2025IJCAI} & \cellcolor{c1!36}\underline{22.95$\pm$0.33} & 0.541$\pm$0.010 & 0.6215 & 0.6535 & 0.6125 & 0.8715 & 0.9022 & 0.9233 & 0.6753 & 0.6556 & 0.6753\\
 IGIB-ISE \cite{zhang2025iterative} & 23.83$\pm$0.26 & 0.572$\pm$0.024 & \cellcolor{c1!36}\underline{0.6340} & 0.6124 & 0.6231 & \cellcolor{c1!36}\underline{0.8949} & 0.9277 & 0.9302 & 0.6827 & \cellcolor{c1!80}\textbf{0.7215} & 0.7009\\
 Meta-MolNet \cite{10436119} & 25.95$\pm$0.36 & 0.596$\pm$0.026 & 0.6019 & 0.6352 & 0.6022 & 0.8536 & 0.8741 & 0.8536 & 0.6571 & 0.6291 & 0.6322\\
 BIT \cite{zhu2025generalist} & 26.35$\pm$0.51 & 0.582$\pm$0.016 & 0.6035 & 0.6273 & 0.5995 & 0.8632 & 0.8752 & 0.8465 & 0.6623 & 0.6587 & 0.6235\\
\midrule
 DisTrans$_{30\%}$ & 24.66$\pm$0.35 & 0.556$\pm$0.021 & 0.6255 & 0.6784 & 0.6322 & 0.8602 & 0.9257 & 0.9198 & 0.7026 & 0.6812 & 0.7232\\
 DisTrans$_{50\%}$ & 23.85$\pm$0.27 & 0.541$\pm$0.029 & 0.6292 & \cellcolor{c1!36}\underline{0.6811} & \cellcolor{c1!36}\underline{0.6437} & 0.8646 & 0.9316 & 0.9367 & \cellcolor{c1!36}\underline{0.7191} & 0.6977 & \cellcolor{c1!36}\underline{0.7332}\\
 DisTrans$_{70\%}$ & \cellcolor{c1!80}\textbf{22.36$\pm$0.25} & \cellcolor{c1!80}\textbf{0.528$\pm$0.026} & \cellcolor{c1!80}\textbf{0.6431} & \cellcolor{c1!80}\textbf{0.7059} & \cellcolor{c1!80}\textbf{0.6524} & 0.8903 & \cellcolor{c1!80}\textbf{0.9636} & \cellcolor{c1!80}\textbf{0.9560} & \cellcolor{c1!80}\textbf{0.7372 }& 0.7155 & \cellcolor{c1!80}\textbf{0.7486}\\
 Imp (\%) & \textcolor{red}{+2.57} & \textcolor{red}{+1.86} & \textcolor{red}{+1.44} & \textcolor{red}{+3.79} & \textcolor{red}{+1.68} & -0.81 & \textcolor{red}{+0.92} & \textcolor{red}{+1.12} & \textcolor{red}{+4.05} & -0.83 & \textcolor{red}{+4.22} \\
\bottomrule 
\end{tabular}
}
}
\end{table*}

\textbf{DisTrans consistently demonstrates the best cross-domain prediction performance in both cross-domain strategies.} As shown in Tables~\ref{IDHT result} and~\ref{CDT result}, DisTrans achieves the best overall performance in IDHT and CDT cross-domain strategies. Compared to IDHT, which focuses on knowledge transfer between datasets within the same domain, CDT involves heterogeneous domain transfer and poses greater challenges to the model. DisTrans consistently outperforms all baseline methods in CDT, with performance improvements exceeding 10\% across multiple evaluation metrics. Compared with molecular cross-domain models (Meta-MolNet and BIT), DisTrans demonstrates significant advantages in IDHT and CDT. These results confirm the effectiveness of the modality-aware heterogeneous cross-domain strategy in DisTrans, which facilitates knowledge transfer between domains and significantly enhances the model’s domain generalization ability.

\textbf{DisTrans achieves efficient cross-domain transfer even when relying on medium-scale target domain data.} As shown in Table~\ref{IDHT result}, DisTrans achieves prediction performance comparable to non-cross-domain models using merely 50\% of the target domain data. Further analysis combining the performance results in Table~\ref{IDHT result} reveals that the performance advantage of DisTrans is particularly pronounced on small-to-medium-sized datasets (\textit{e.g.}, DDInter2.0 and TCMM). Specifically, DisTrans trained with 50\% of the target domain data outperforms all baseline methods across multiple evaluation metrics in these datasets. DisTrans surpasses the HHI prediction performance of the comparative models with only 30\% of the target-domain data in CDT. These results indicate that DisTrans can achieve efficient cross-domain transfer and molecular representation learning with low target domain data dependence.

\subsection{Data Transfer Preference Analysis (\textbf{RQ2})}
\label{Data Transfer Preference Analysis2}
\begin{figure*}[t]
\centering
\includegraphics[width=\textwidth]{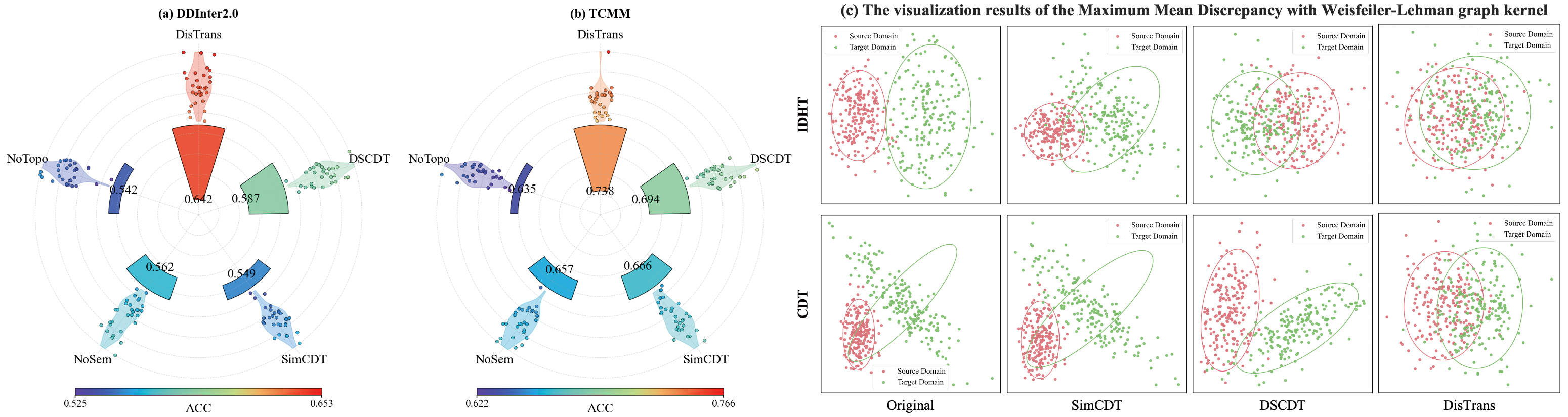} 
\caption{The performance and representation differences of different transfer strategies. (a) and (b) present the ACC in IDHT.}
\label{rq2}
\end{figure*}

We investigate the cross-domain transfer preferences of different types of molecular data by introducing multiple cross-domain strategies. SimCDT and DSCDT aim to maximize representational similarity and domain-specificity between domains, respectively. NoTopo and NoSem only consider semantic consistency and structural domain-specificity in isolation. The detailed experimental settings are provided in Appendix \ref{Data Transfer Preference Analysis}; the results are shown in Fig.~\ref{rq2}, Fig.~\ref{rq3-supp} and Table \ref{estimation results}.

\textbf{Distinct cross-domain transfer preferences are observed across different molecular data types.} As shown in Figs.~\ref{rq2} and~\ref{rq3-supp}, cross-domain learning mechanisms that disregard data-type heterogeneity (SimCDT and DSCDT) exhibit varying degrees of degradation in predictive performance and representation consistency between domains. These results suggest that molecular topological structures and image semantics exhibit significantly different transfer requirements in CDMRL. Additionally, strategies that separately transfer single modality (NoSem and NoTopo) result in approximately 6\% to 8\% performance loss, further confirming the synergistic role of topological representations and semantic information in CDMRL. In contrast, DisTrans achieves the best cross-domain prediction performance and molecular representation by explicitly accounting for the differences in transfer requirements between molecular structures and images.

\textbf{Molecular structures prefer domain-specific transfer and have a more significant impact on CDMRL.} The comparison between DisTrans and SimCDT reveals that enforcing topological representation consistency suppresses cross-domain transfer performance and causes substantial fluctuations. These phenomena indicate that domain-specific topological representations are more conducive to optimizing CDMRL performance. The experimental results in Fig.~\ref{rq3-supp} also support this conclusion.
Moreover, the predictive performance of NoTopo on both datasets is significantly lower than that of NoSem, and the performance comparison between SimCDT and DSCDT exhibits the same trend. The inter-domain mutual information estimates in Table \ref{estimation results} indicate that structural mutual information shifts are the primary factor influencing molecular cross-domain representations. This collectively demonstrates that structural domain-specific representation is the dominant factor in CDMRL, consistent with the conclusion in Theorem 1.

\begin{table*}[t]
\caption{The prediction results of DisTrans and baseline methods in Cross-Domain Transfer (CDT).}
\label{CDT result}
\centering
{\fontsize{9pt}{10pt}\selectfont
\setlength{\tabcolsep}{0.45mm}{
\begin{tabular}{l|ccc|ccc|ccc|ccc|ccc}
\toprule
\multirow{3}{*}{Model} & \multicolumn{9}{c|}{DDI ($\mathcal{D}_s$: ITCM)} & \multicolumn{6}{c}{HHI ($\mathcal{D}_s$: ZhangDDI)}\\
 \cline{2-16}
& \multicolumn{3}{c|}{ZhangDDI} & \multicolumn{3}{c|}{DDInter2.0} & \multicolumn{3}{c|}{DrugMap2024} & \multicolumn{3}{c|}{TCMM}& \multicolumn{3}{c}{ITCM}\\
& ACC & AUROC & AUPR & ACC & AUROC & AUPR & ACC & AUROC & AUPR & ACC & AUROC & AUPR& ACC & AUROC & AUPR\\ 
 \midrule
 GCN \cite{gcn} & 0.4991 & 0.4591 & 0.4937 & 0.4757 & 0.4667 & 0.4373 & 0.5454 & 0.5469 & 0.5714 & 0.4018 & 0.1536 & 0.4336 & 0.5952 & 0.4126 & 0.5342\\
 GAT \cite{gat} & 0.5006 & 0.5292 & 0.5061 & 0.4713 & 0.4322 & 0.4221 & 0.5017 & 0.5039 & 0.5411 & 0.4072 & 0.1388 & 0.4385 & 0.6007 & 0.4112 & 0.5375 \\
 MPNN \cite{mpnn} & 0.5094 & 0.4987 & 0.4853 & 0.4981 & 0.4821 & 0.4439 & 0.5477 & 0.5495 & 0.5678 & 0.4842 & 0.1581 & 0.4794 & 0.5946 & 0.4133 & 0.5354 \\
 GINE \cite{Hu2020Strategies} & 0.5125 & 0.5272 & 0.5137 & 0.4728 & 0.4386 & 0.5245 & 0.5325 & 0.5335 & 0.5488 & 0.4618 & 0.2316 & 0.4504 & 0.5978 & 0.4190 & 0.5358 \\
 SSI-DDI \cite{ssiddi} & 0.5203 & 0.5315 & 0.5333 & 0.5496 & 0.5646 & 0.5747 & 0.5325 & 0.4523 & 0.5055 & 0.5582 & 0.4866 & 0.5468 & 0.5799 & 0.5852 & 0.5834 \\
 SA-DDI \cite{saddi} & 0.3684 & 0.3118 & 0.3765 & 0.5077 & 0.5636 & 0.5647 & 0.4915 & 0.4738 & 0.4988 & 0.5176 & 0.4687 & 0.5564 & 0.5666 & 0.5743 & 0.5687 \\
 DSN-DDI \cite{dsn} & 0.5107 & 0.5108 & 0.5447 & 0.4691 & 0.4678 & 0.4811 & \cellcolor{c1!36}\underline{0.6328} & 0.6271 & 0.6373 & 0.4746 & 0.4788 & 0.5064 & 0.5662 & 0.5336 & 0.5648 \\
 CIGIN \cite{cigin} & 0.5122 & 0.5058 & 0.5146 & 0.5273 & 0.5296 & 0.5663 & 0.5673 & 0.5748 & 0.5859 & 0.4722 & 0.4242 & 0.4652 & 0.5316 & 0.5426 & 0.5512 \\
 MIRACLE \cite{miracle} & 0.4928 & 0.4975 & 0.5384 & 0.5138 & 0.4152 & 0.5475 & 0.4873 & 0.3748 & 0.4054 & 0.5038 & 0.4322 & 0.5021 & 0.5347 & 0.5607 & 0.5614 \\
 CMRL \cite{lee2023shiftKDD} & 0.5159 & 0.5055 & 0.5541 & 0.5346 & 0.5745 & 0.5794 & 0.5891 & 0.6007 & 0.5358 & 0.5217 & 0.5416 & 0.5905 & 0.6038 & 0.6164 & 0.6162 \\
 CGIB \cite{lee2023cgibICML} & 0.5231 & \cellcolor{c1!36}\underline{0.5356} & 0.6082 & 0.5509 & \cellcolor{c1!36}\underline{0.5751} & 0.5646 & 0.5942 & \cellcolor{c1!36}\underline{0.6326} & 0.6134 & 0.5223 & 0.5567 & 0.5896 & 0.6243 & 0.6304 & 0.6161 \\
 MMGNN \cite{du2024mmgnnIJCAI} & 0.5037 & 0.5245 & 0.6133 & 0.5034 & 0.5112 & 0.5206 & 0.5244 & 0.5244 & 0.5643 & 0.5246 & 0.4978 & 0.5916 & 0.5776 & 0.5911 & 0.5961 \\
 CausalGIB \cite{zhang2025IJCAI} & 0.5232 & 0.5102 & 0.6251 & 0.5096 & 0.5125 & 0.5351 & 0.5685 & 0.5881 & 0.5633 & 0.5363 & 0.4862 & 0.5921 & 0.5636 & 0.6112 & 0.5896 \\
 IGIB-ISE \cite{zhang2025iterative} & 0.5434 & 0.5269 & 0.6328 & 0.5222 & 0.5126 & 0.5631 & 0.6003 & 0.6136 & 0.6103 & 0.5425 & 0.4796 & 0.5914 & 0.5485 & 0.6016 & 0.5866 \\
 Meta-MolNet \cite{10436119} & 0.5036 & 0.5256 & 0.6521 & 0.5195 & 0.5232 & 0.5862 & 0.6125 & 0.6068 & 0.5985 & 0.5326 & 0.5022 & 0.5865 & 0.5535 & 0.6022 & 0.6112\\
 BIT \cite{zhu2025generalist} & 0.4963 & 0.5116 & 0.6495 & 0.5336 & 0.5163 & 0.5962 & 0.6097 & 0.6003 & 0.5369 & 0.5123 & 0.5895 & 0.5896 & 0.6101 & 0.5962 & 0.6026\\
 \midrule
 DisTrans$_{30\%}$ & 0.4966 & 0.5012 & 0.6635 & 0.5435 & 0.5526 & 0.6215 & 0.6210 & 0.5993 & 0.6102 & 0.5636 & \cellcolor{c1!36}\underline{0.5991} & 0.6001 & 0.6015 & 0.63148 & 0.6325 \\
 DisTrans$_{50\%}$ & \cellcolor{c1!36}\underline{0.5526} & 0.5222 & \cellcolor{c1!36}\underline{0.6678} & \cellcolor{c1!36}\underline{0.5547} & 0.5658 & \cellcolor{c1!36}\underline{0.6297} & 0.6236 & 0.6241 & \cellcolor{c1!36}\underline{0.6293} & \cellcolor{c1!36}\underline{0.5986} & 0.5229 & \cellcolor{c1!36}\underline{0.6036} & \cellcolor{c1!36}\underline{0.6291} & \cellcolor{c1!36}\underline{0.6453} & \cellcolor{c1!36}\underline{0.6526} \\
 DisTrans$_{70\%}$ & \cellcolor{c1!80}\textbf{0.5974} & \cellcolor{c1!80}\textbf{0.5892} & \cellcolor{c1!80}\textbf{0.7052} & \cellcolor{c1!80}\textbf{0.6040} & \cellcolor{c1!80}\textbf{0.6276} & \cellcolor{c1!80}\textbf{0.6543} & \cellcolor{c1!80}\textbf{0.6473} & \cellcolor{c1!80}\textbf{0.6763} & \cellcolor{c1!80}\textbf{0.6856} & \cellcolor{c1!80}\textbf{0.6245} & \cellcolor{c1!80}\textbf{0.6078} & \cellcolor{c1!80}\textbf{0.6561} & \cellcolor{c1!80}\textbf{0.6498} & \cellcolor{c1!80}\textbf{0.6565} & \cellcolor{c1!80}\textbf{0.6739}\\
 Imp (\%) & \textcolor{red}{+9.94} & \textcolor{red}{+8.83} & \textcolor{red}{+11.44} & \textcolor{red}{+9.64} & \textcolor{red}{+9.13} & \textcolor{red}{+12.93} & \textcolor{red}{+2.29} & \textcolor{red}{+6.59} & \textcolor{red}{+12.39} & \textcolor{red}{+11.88} & \textcolor{red}{+9.18} & \textcolor{red}{+10.90} & \textcolor{red}{+4.08} & \textcolor{red}{+4.14} & \textcolor{red}{+9.36} \\
\bottomrule
\end{tabular}
}
}
\end{table*}

\begin{table}[htpb]
  \caption{The estimation results of mutual information under different cross-domain settings.}
  \label{estimation results}
  \centering
  {\fontsize{8pt}{10pt}\selectfont
\setlength{\tabcolsep}{0.5mm}{
\begin{tabular}{l|cc|cc}
\toprule
\multirow{2}{*}{}& \multicolumn{2}{c|}{IDHT} & \multicolumn{2}{c}{CDT}\\
 \cline{2-5}
& DDI(DrugMap2024) & HHI(TCMM) & DDI(DDInter2.0) & HHI(ITCM)\\
 \midrule
 $\Delta \mathcal{I}$ & 4.791 bits & 3.734 bits & 8.481 bits & 7.772 bits\\
 $\Delta \mathcal{I}^{(g)}$ & 3.265 bits & 2.532 bits & 5.256 bits & 4.618 bits\\
 $\Delta \mathcal{I}^{(i|g)}$ & 1.526 bits & 1.202 bits & 3.225 bits & 3.154 bits \\
\bottomrule 
\end{tabular}
}
}
\end{table}

\subsection{Transfer Strategy Transferability Analysis (\textbf{RQ3})}
\label{Transferability Analysis of Cross-Domain Strategies}
\begin{figure}[t]
\centering
\includegraphics[width=\columnwidth]{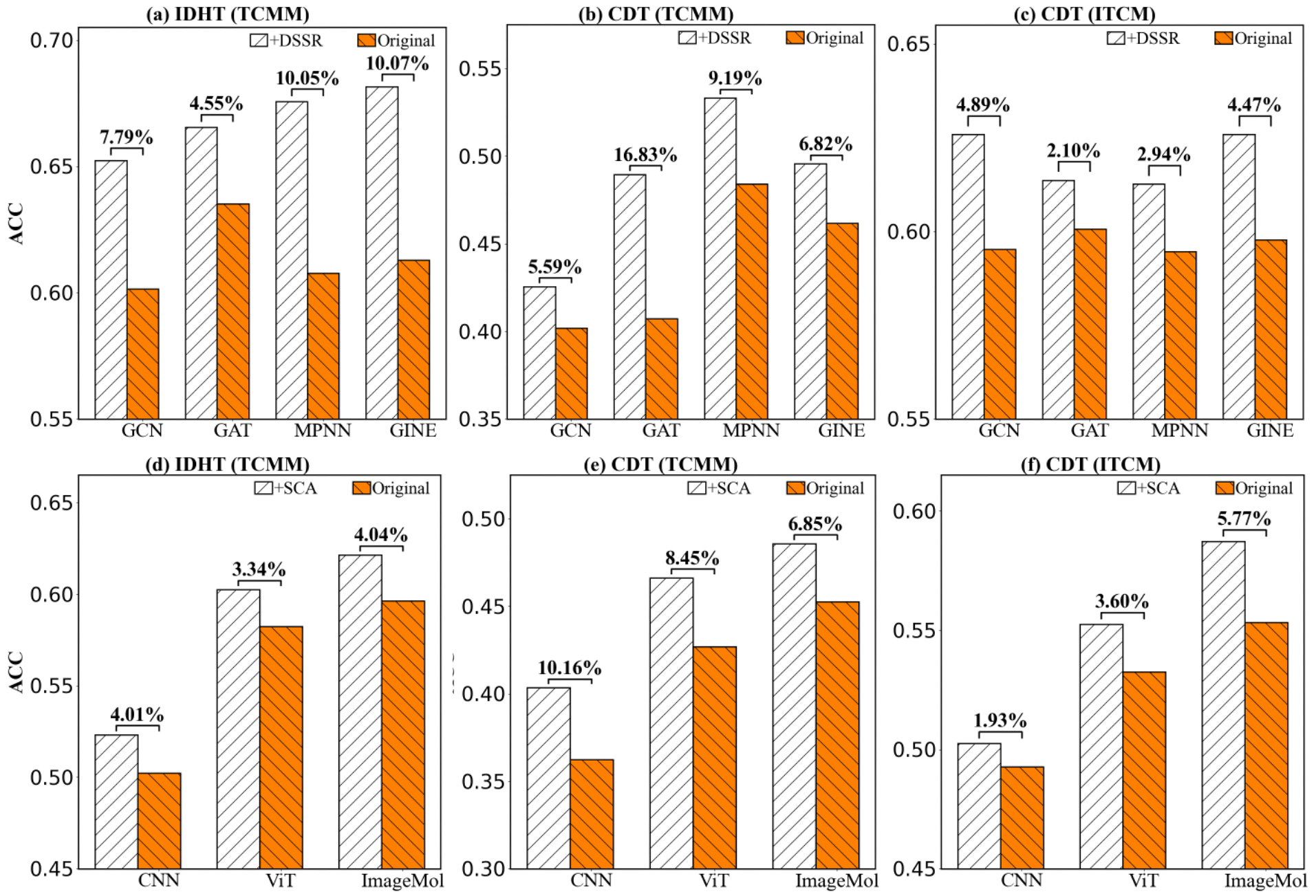} 
\caption{The cross-domain adaptability prediction performance of DisDAT with different benchmark models.}
\label{figrq3}
\end{figure}

We integrated DSSR and SCA into the graph benchmark models (GCN, GAT, MPNN, GINE), visual benchmark models (CNN and ViT), and molecular image representation model (ImageMol) to evaluate the adaptability and generality of the DisDAT in CDMRL. The experimental results are shown in Fig \ref{figrq3}, where SCA is semantic similarity alignment without $\mathbb{M}$.

\textbf{The cross-domain mechanism leveraging structural-semantic transfer discrepancy demonstrates strong compatibility with benchmark models.} As shown in Fig \ref{figrq3}, DSSR and SCA consistently enhance the predictive performance of benchmark models in IDHT and CDT transfer strategies, with particularly significant performance improvements in the more challenging CDT strategy. These results indicate that the structural-semantic discrepancy transfer mechanism exhibits high model adaptability and transfer generality. In CDT, domain-specific structural representations deliver performance improvements exceeding 5\% across multiple evaluation metrics. These results validate the effectiveness of domain-dependent topological features in CDMRL.

\begin{figure}[htpb]
\centering
\includegraphics[width=\columnwidth]{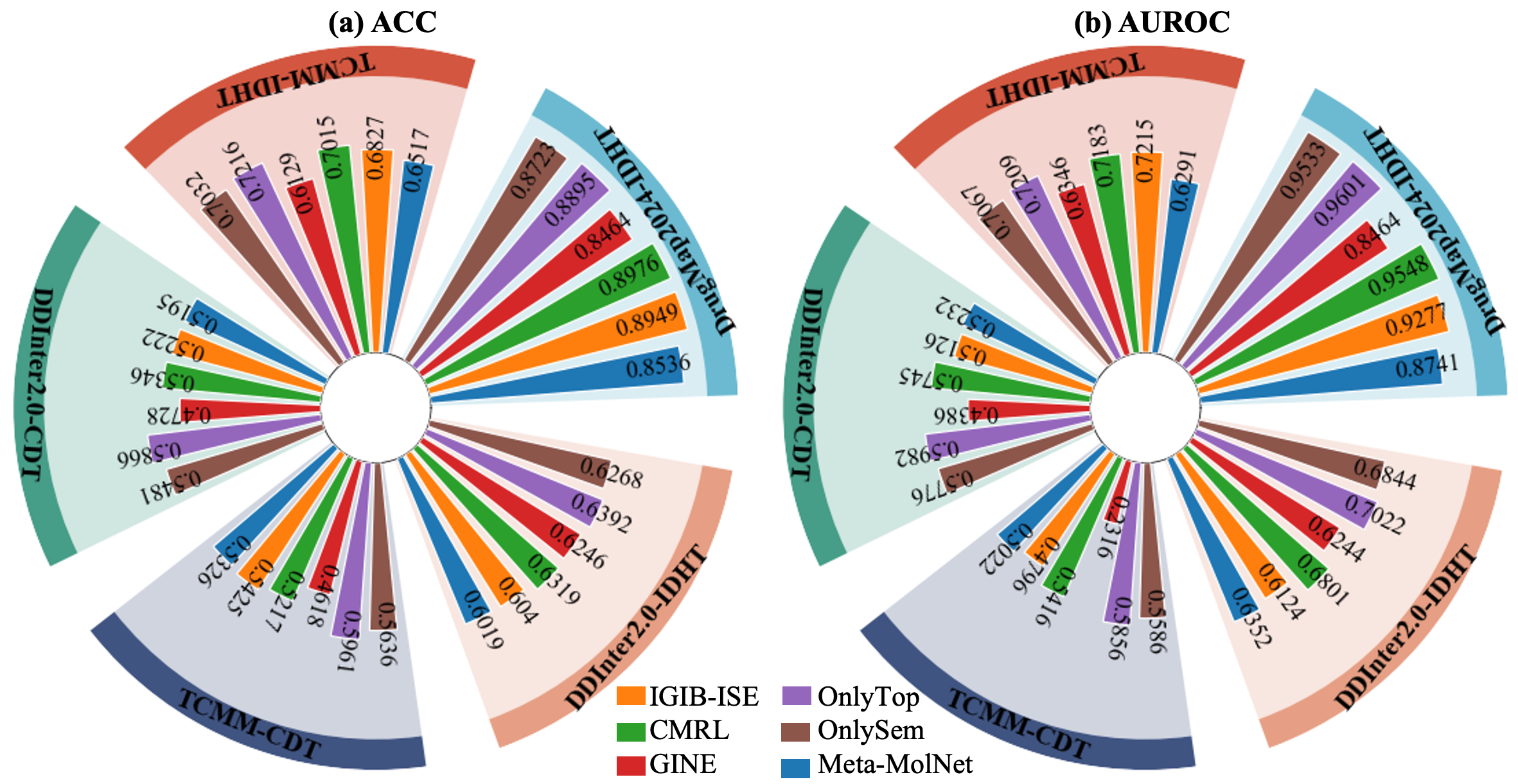} 
\caption{The predictive Performance of OnlyTop, OnlySem, and Comparison Models in IDHT and CDT}
\label{ABAA}
\end{figure}

Additionally, we investigate the impact of transferring single-modality information on CDMRL, including only differential transfer topological representations (OnlyTop) and only consistency-based transfer semantic information (OnlySem). Unlike the NoTopo and NoSem configurations in Section \ref{Data Transfer Preference Analysis2}, OnlyTop and OnlySem focus solely on single-modal molecular representations and their cross-domain transfer. The results are shown in Fig. \ref{ABAA}.

Domain-specific topological representations are more beneficial for CDMRL. As shown in Fig. \ref{ABAA}, OnlyTop, which considers only domain-specific molecular representations, achieves the best overall predictive performance in the IDHT and CDT settings. Notably, its advantage is more pronounced in the CDT setting, outperforming the comparison method by over 5\%. This improvement arises because OnlyTop emphasizes learning cross-domain topological structural discrepancies, thereby strengthening the domain dependency of molecular representations.
Although semantic-consistency representations also enhance CDMRL performance, their gains are substantially smaller than those of OnlyTop. These results indicate that domain-specific molecular representations play a more critical role in CDMRL. This observation is also consistent with the conclusion of Theorem 1 and the results reported in Table \ref{estimation results}, which jointly suggest that domain-specific structural representations are more advantageous for CDMRL.

\subsection{Visualization Analysis (RQ4)}
\begin{figure}[t]
\centering
\includegraphics[width=\columnwidth]{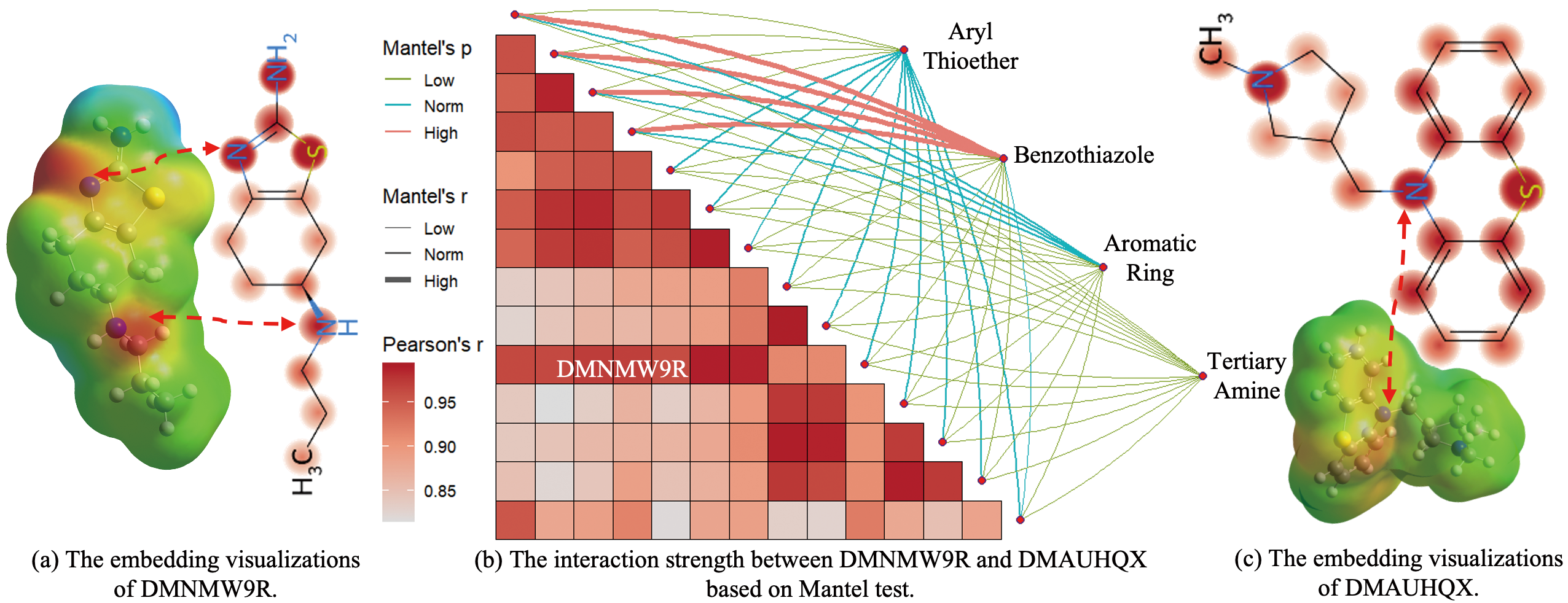}
\caption{The visualization results of molecular representation and substructure interaction strength in CDT.}
\label{vis}
\end{figure}

In this section, we analyze cross-domain representational discrepancies across different molecular types and evaluate the effectiveness of DisTrans in molecular representation by visualizing the molecular representations. 
The molecular representations and substructure interaction strength results are shown in Fig.~\ref{vis}.

\textbf{DisTrans demonstrates excellent capabilities in domain-specific structural representation and core substructure learning in CDMRL.} Figs.~\ref{vis}(a) and~(c) show the correspondence between molecular electrostatic potentials and the structural representations learned by DisTrans. Taking DMNMW9R as an example, we observe significant representational differences between carbon and sulfur atoms, with DisTrans focusing more on the representation of amino groups. This observation is consistent with the chemical electrostatic potential distribution, indicating that the cross-domain representations learned by DisTrans are chemically interpretable. Furthermore, the Mantel test results of substructure interactions in Fig.~\ref{vis}(b) show that DisTrans effectively identifies the Benzothiazole (red lines) with strong interactions with DMNMW9R. This is reflected in the progressively intensified interaction strengths among core substructures, while interactions among other substructures gradually weakened. This phenomenon is consistent with the domain knowledge from the chemically induced fit theory that molecular interactions predominantly occur between core substructures.
Overall, the cross-domain strategy based on structural-semantic transfer discrepancy endows DisTrans with significant advantages in molecular cross-domain representation.

\section{Conclusion}
This paper proposes a molecular cross-domain representation framework benefiting from chemical structure-activity analysis. DisTrans integrates modality-aware heterogeneous cross-domain strategy into domain adversarial training, enabling cross-domain adaptive molecular representation by differentially transferring modal representation information. Experimental results demonstrate that DisTrans performs superior in two cross-domain strategies while reducing molecular representation discrepancies between domains.
Future research will focus on weakly supervised cross-domain molecular learning to enhance the stability and effectiveness of differential transfer mechanisms in scenarios with scarce target-domain annotations or source-domain invisibility.

\section*{Acknowledgement}
This work was supported by the National Natural Science Foundation of China (No. 62472332 and No. 62276196), the Hainan Provincial Natural Science Foundation of China (No. 526MS0269), the Basic Project for Universities from the Educational Department of Liaoning Province (No. LJ242511258004), the 111 Project (No. D23006).

\bibliographystyle{ACM-Reference-Format}
\bibliography{sample-base}

\appendix

\setcounter{table}{0}
\renewcommand{\thetable}{S\arabic{table}}
\setcounter{figure}{0}
\renewcommand{\thefigure}{S\arabic{figure}}

\section{The Proof of Theorem 1}
\label{The Proof of Theorem 1}
For any model representation $Z=f_\theta(M)$, the prediction errors in the source and target domains are defined as:
\begin{equation}
\begin{aligned}
    &\varepsilon_s=\mathbb{E}_{(z,y) \sim D_s}[-\log P(y|z)]\\
    &\varepsilon_t=\mathbb{E}_{(z,y) \sim D_t}[-\log P(y|z)]
    \label{varepsilon}
\end{aligned}
\end{equation}
Where $P(\cdot)$ is a conditional probability function.

According to the chain decomposition of multivariate mutual information $H(Y|Z)=H(Y)-\mathcal{I}(Z;Y)$, $\Delta \varepsilon$ can be further expressed as:
\begin{equation}
\begin{aligned}
    \Delta \varepsilon &\!\!=\!\!\underbrace{\mathbb{E}_{(z,y) \sim D_t} \![\!-\! \log P(y|z)\!]}_{\Delta H(Y_t|Z_t)}\!\!-\!\!\underbrace{\mathbb{E}_{(z,y) \sim D_s}\! [\!- \!\log P(y|z)\!]}_{\Delta H(Y_s|Z_s)}\\
    &\left.
    \begin{aligned}
    \!=\!\mathbb{E}_{y \sim D_t}\![- \!\log P(y)]\!\!-\!\!\mathbb{E}_{y \sim D_s}\![-\! \log P(y)]
    \end{aligned}
    \right \}\! \Delta H(Y)\\
    & \left.
    \begin{aligned}
    +\mathbb{E}_{(Z^{(g)},y) \sim D_s}\left[ \log \frac{P(y|Z^{(g)})}{P(y)} \right]\\
    -\mathbb{E}_{(Z^{(g)},y) \sim D_t}\left[ \log \frac{P(y|Z^{(g)})}{P(y)} \right]
    \end{aligned}
    \right \} \Delta \mathcal{I}^{(g)}\\
    & \left.
    \begin{aligned}
    +\mathbb{E}_{(Z^{(g)}\!,Z^{(i)}\!,y) \sim D_s}\!\!\left[ \log \frac{P(y|Z^{(g)}\!,\!Z^{(i)})}{P(y|Z^{(g)})} \!\right]\\
    -\mathbb{E}_{(Z^{(g)}\!,Z^{(i)}\!,y) \sim D_t}\!\!\left[ \log \frac{P(y|Z^{(g)}\!,\!Z^{(i)})}{P(y|Z^{(g)})} \!\right]
    \end{aligned}
    \right \} \Delta \mathcal{I}^{(i|g)}\\
    \label{Delta varepsilon new}
\end{aligned}
\end{equation}
Here,  $\Delta \mathcal{I}$, defined as the difference between inter-domain KL divergences, reflects domain-label dependencies, not mere distributional shifts. $\Delta \mathcal{I}^{(g)}$ and $\Delta \mathcal{I}^{(i|g)}$ represent the cross-domain structure-activity shifts of graph structures and image semantics, respectively, which are calculated based on the molecular representation $Z=(Z^{(g)},Z^{(i)})$ and the mutual information chain rule $\mathcal{I}(Z;Y)=\mathcal{I}(Z^{(g)};Y)+\mathcal{I}(Z^{(i)};Y|Z^{(g)})$.

The DDI and HHI prediction tasks addressed in this paper are both inherently classification tasks. Since the task label distributions between the source domain and the target domain are consistent in this study, we have $\Delta H(Y)=0$. Therefore, there exists a real number $\xi >0$ such that:
\begin{equation}
   \Delta\varepsilon \geq \xi (\Delta\mathcal{I}^{(g)}+\Delta\mathcal{I}^{(i|g)}) 
   \label{eq5}
\end{equation}

Next, we proceed to demonstrate that $\Delta\mathcal{I}^{(g)} \gg \Delta\mathcal{I}^{(i|g)}$. First, $\Delta \mathcal{I}^{(g)}$ and $\mathcal{I}^{(i|g)}$ in Eq (\ref{Delta varepsilon new}) are simplified as:
\begin{equation}
    \Delta \mathcal{I}^{(g)}=\mathcal{I}(Z^{(g)}_{s};Y_s)-\mathcal{I}(Z^{(g)}_{t};Y_t)\\
    \label{Delta Ig}
\end{equation}
\begin{equation}
    \Delta \mathcal{I}^{(i|g)}=\mathcal{I}(Z^{(i)}_{s};Y_s|Z^{(g)}_{s})-\mathcal{I}(Z^{(i)}_{t};Y_t|Z^{(g)}_{t})\\
    \label{Delta Ii}
\end{equation}
Where
\begin{equation}
    \mathcal{I}(Z^{(g)};Y)=D_{KL}(P(Z^{(g)},Y)||P(Z^{(g)})P(Y))
\end{equation}
\begin{equation}
\begin{aligned}
    \mathcal{I}(Z^{(i)};Y|Z^{(g)})\!=\!\mathbb{E}_{Z^{(g)}} \! [&D_{KL}\!(P(Z^{(i)},Y|Z^{(g)})||\\
    &P(Z^{(i)}|Z^{(g)})P(Y|Z^{(g)}) )  ]
\end{aligned}
\end{equation}
$D_{KL}$ denotes the Kullback–Leibler (KL) divergence.

\textbf{Structural Mutual Information:} Due to the inconsistent molecular structure distribution between the source and target domains, $P(Z^{(g)}, Y)$ and $P(Z^{(g)})$ match poorly in the target domain. In contrast, $\mathcal{I}(Z^{(g)}_{s};Y_s)$ can be approximately saturated in the source domain. Accordingly, we obtain:
\begin{equation}
    \Delta \mathcal{I}^{(g)}=D_{KL}^s-D_{KL}^t \gg 0
    \label{sturcture}
\end{equation}

\textbf{Image Mutual Information:} Since visual information primarily provides semantically stable features such as functional group types, and $Z^{(i)}$ serves as conditional supplementary information for $Z^{(g)}$, its representations exhibit robustness across domains. Thus,
\begin{equation}
    \mathcal{I}(Z^{(i)}_{s};Y_s|Z^{(g)}_{s}) \leq \mathcal{I}(Z^{(i)}_{s};Y_s)
\end{equation}

Therefore,
\begin{equation}
     \Delta \mathcal{I}^{(i|g)}\!=\!\mathbb{E}_{Z^{(g)}}\!\! \left [ D_{KL}^s(Z^{(i)})\!-\!D_{KL}^t(Z^{(i)})\right ]\! \ll \!\Delta \mathcal{I}^{(g)}
    \label{image}
\end{equation}

Combining Eq (\ref{eq5}) with Eq (\ref{image}), we prove that
\begin{equation}
    \Delta\varepsilon \geq \xi (\Delta\mathcal{I}^{(g)}+\Delta\mathcal{I}^{(i|g)}),\quad \Delta\mathcal{I}^{(g)} \gg \Delta\mathcal{I}^{(i|g)}
\end{equation}

Theory 1 is proven.

In summary, we provide a mathematical decomposition explanation for the domain-specific assertion that inter-domain structural disparities induce more severe information bias.

\section{Experimental Results and Analysis}
\label{Experimental Results and Analysis}
\subsection{Imp}
\label{Overall Performance Analysis}
Imp denotes the performance improvement magnitude of DisTrans.
\begin{equation}
    \text{Imp}=\frac{\text{DisTrans}-\text{Best \, Baseline}}{\text{Best \, Baseline}} \times 100\%
    \label{imp}
\end{equation}

\begin{figure*}[htpb]
\centering
\includegraphics[width=\textwidth]{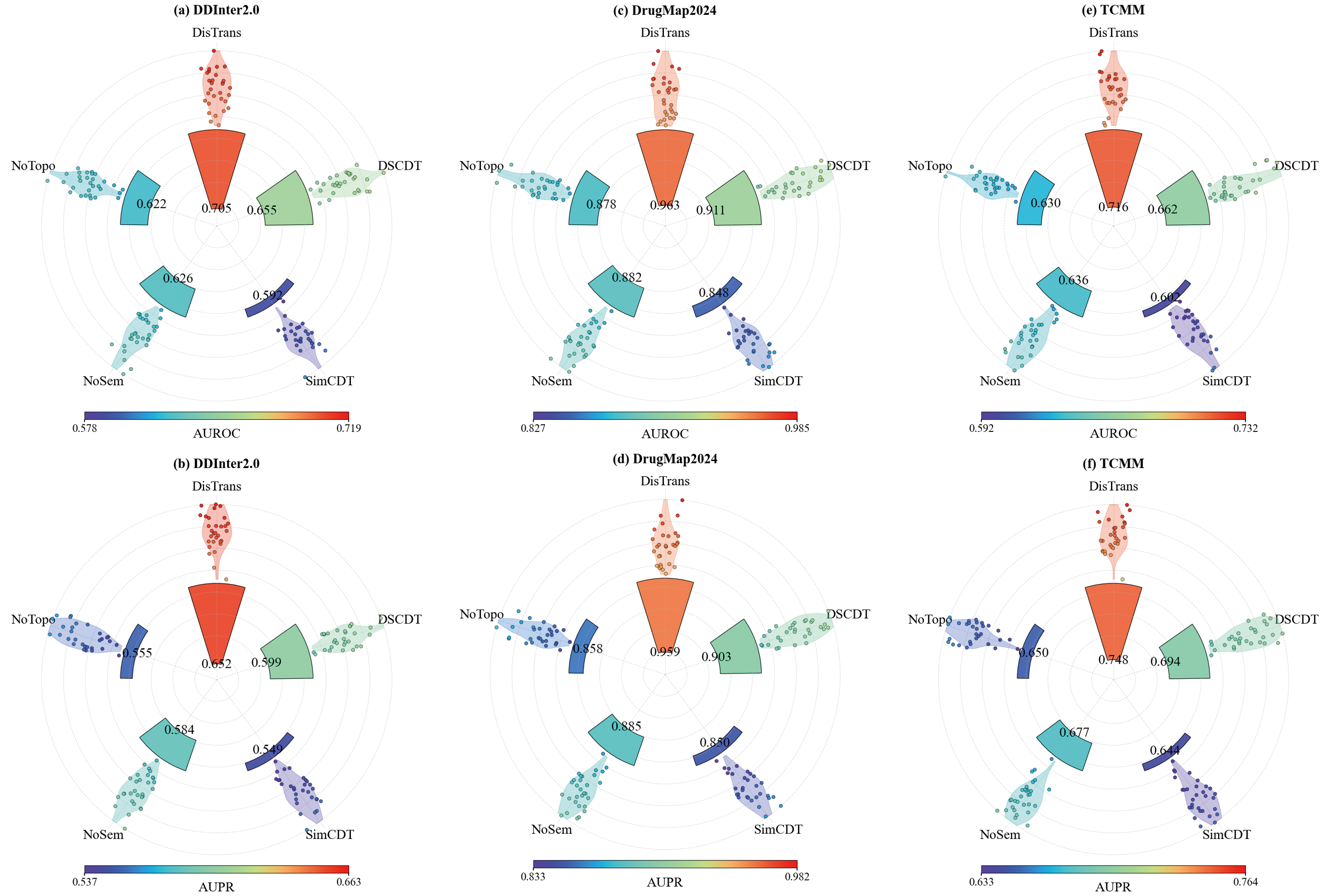}
\caption{The prediction results of different cross-domain mechanisms in IDHT.}
\label{rq3-supp}
\end{figure*}

\subsection{Data Transfer Preference Analysis}
\label{Data Transfer Preference Analysis}
We investigate the cross-domain transfer preferences of different types of molecular data by introducing multiple cross-domain strategies. The experimental settings are as follows:
\begin{itemize}
    \item \textbf{Similarity-based Cross-Domain Transfer (SimCDT):} SimCDT aims to maximize the representational similarity between the source and target domains to achieve alignment of inter-domain embeddings. In the experimental setup, SimDCT employs SCA to facilitate the cross-domain transfer of functional group semantic information and molecular topological structure representation.
    \item \textbf{Domain-specificity-based Cross-Domain Transfer (DSCDT):} DSCDT maximizes inter-domain differences to generate domain-specific representations for the target domain. In the experimental setup, DSCDT utilizes DSSR to transfer functional group semantic information and molecular topological representations between domains.
    \item \textbf{Topology Representation not Transferred (NoTopo):} NoTopo relies solely on semantic consistency alignment to transfer prior knowledge representations across domains, without involving the transfer of topological representations. In the experimental setup, NoTopo leverages only SCA to transfer and generate molecular functional group semantic information, while the molecular topological representations are represented by $Z^{(g)}$ generated by TRegCross.
    \item \textbf{Semantic Information not Transferred (NoSem):} In this case, domain-specific representations of target domain topologies are generated exclusively through domain-specificity enhancement, without transferring or aligning semantic information. In the experimental setup, NoSem exclusively employs DSSR to transfer and generate molecular topological representations, with functional-group semantic information represented by $Z^{(i)}$ generated by TRegCross.
\end{itemize}

Fig.~\ref{rq3-supp} shows the prediction performance differences of different cross-domain mechanisms in IDHT and CDT, where  Figs.~\ref{rq3-supp}(a)-(c) are AUROC, and Figs.~\ref{rq3-supp}(d)-(f) are AUPR. Strategies that neglect data types or transfer only single-type data both negatively impact cross-domain performance, with certain metrics showing performance declines exceeding 10\%. By comparing the results of DisTrans with SimCDT and DSCDT, we observe that image semantics and structural topology exhibit distinct transfer preferences in cross-domain learning. Specifically, image semantics tend to favour inter-domain consistency alignment, whereas structural representations require emphasis on domain specificity. These experimental observations are consistent with the design rationale of cross-domain strategy based on structural-semantic transfer discrepancy.

\end{document}